# Capturing Complex Spatial-Temporal Dependencies in Traffic Forecasting: A Self-Attention Approach

Chenghong Zheng[‡]
School of Mathematics and Statistics
Guangdong University of Technology
Guangzhou, China
3121006962@mail2.gdut.edu.cn

Zongyin Deng[‡]
Department of Computational Mathematics and Control
Shenzhen MSU-BIT University
Shenzhen, China
l1120210006@smbu.edu.cn

[‡] These authors contributed equally to this work.

Cheng Liu, Simin Xiong, Deshi Di, Guanyao Li*
Guangzhou Urban Planning & Design Survey Research Institute Co., Ltd.
Guangzhou, China
{liucheng. xiongsimin, deshi.di}@gzpi.com.cn
* Corresponding author: gyli@gzpi.com.cn

*Abstract*— **We study the problem of traffic forecasting, aiming to predict the inflow and outflow of a region in the subsequent time slot. The problem is complex due to the intricate spatial and temporal interdependence among regions. Prior works study the spatial and temporal dependency in a decouple manner, failing to capture their joint effect. In this work, we propose ST-SAM, a novel and efficient Spatial-Temporal Self-Attention Model for traffic forecasting. ST-SAM uses a region embedding layer to learn time-specific embedding from traffic data for regions. Then, it employs a spatial-temporal dependency learning module based on self-attention mechanism to capture the joint spatial-temporal dependency for both nearby and faraway regions. ST-SAM entirely relies on self-attention to capture both local and global spatial-temporal correlations, which make it effective and efficient. Extensive experiments on two real world datasets show that ST-SAM is substantially more accurate and efficient than the state-of-the-art approaches (with an average improvement of up to 15% on RMSE, 17% on MAPE, and 32 times on training time in our experiments).**

## I. INTRODUCTION

Given the historical traffic flow data, the task of traffic flow forecasting aims to predict the number of objects (e.g., people, vehicles, goods, etc.) arriving at and departing from a region in the next time interval. Traffic flow forecasting has diverse applications in the domains of traffic management, emergency response, urban planning, and other related areas [1]. For example, accurate forecasts help authorities manage traffic flow efficiently by optimizing signal timings, identifying congestion-prone areas, and implementing strategies to alleviate traffic jams. Moreover, predicting traffic flow is vital for emergency services to plan routes, respond promptly to incidents, and ensure the safety and well-being of individuals during critical situations.

However, traffic flow forecasting presents challenges due to the intricate spatial-temporal dependencies among various regions. It has been shown that a region's traffic may have strong spatial-temporal correlation with its neighborhood regions or even distant ones. Moreover, the spatial and temporal patterns may not be independent. This is the case when the traffic of a region is related to, or affected by, that of another region with some temporal lag. Moreover, such spatial correlation between regions may be time-dependent. An example is road condition. When a hot spot develops in an area, often other areas would experience an increase in traffic some time later.

Many studies have been conducted on traffic forecasting by considering spatial and temporal dependency between regions [2-6]. Convolutional Neural Networks (CNNs) and Graph Convolutional Networks (GCNs) are extensively employed for capturing spatial dependencies, whereas Recurrent Neural Networks (RNNs) and their variants like LSTM and GRU are typically utilized for learning temporal dependencies. However, both CNN and GCN tend to capture spatial dependency from nearby regions, overlooking the correlation with distant regions. Moreover, RNNs are inherently sequential in nature, which limits parallelization and can result in slower training times on large datasets. Furthermore, a common trend in these works is the separate consideration of spatial and temporal dependencies, neglecting their joint impact on traffic flow. This gap leaves room for enhancing the precision of traffic forecasting methodologies. Furthermore, the incremental improvements in current studies rely heavily on the extensive training resources. While the model remains static post-training, this situation escalates the costs linked to hyperparameter tuning during model training and deploying large-scale deep learning models.

To tackle the limitations, we propose an efficient **S**patial-**T**emporal **S**elf-**A**ttention **M**odel (ST-SAM) to predict traffic flow in the next time slot (Fig. 1). ST-SAM is entirely relying on self-attention to capture spatial and temporal dependency in a joint manner. With its self-attention mechanisms, ST-SAM is more effective to capture the complicated spatial-temporal correlations between regions, no matter they are nearby or distant. Furthermore, it can be trained in parallel and hence is easy and fast to deploy.

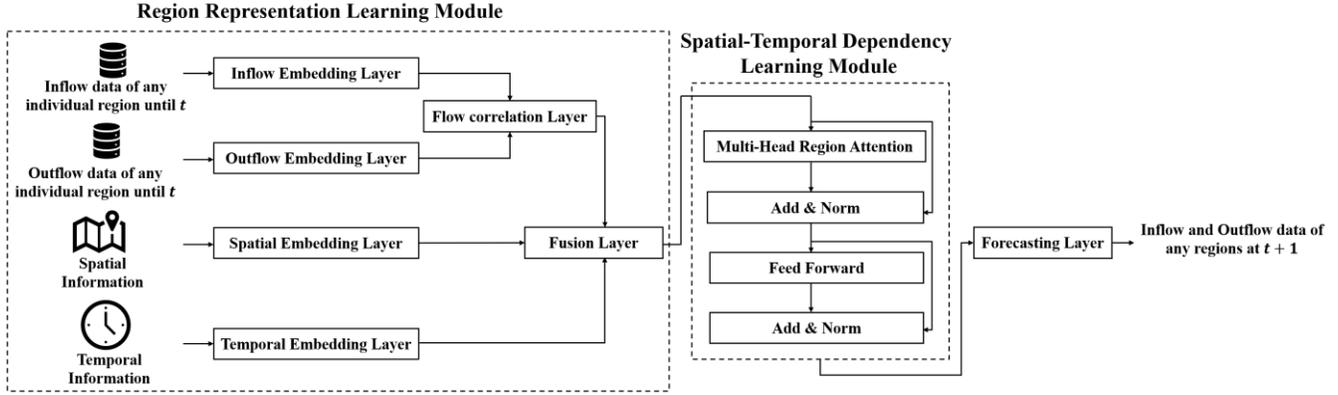

Figure 1. Overview of ST-SAM

In ST-SAM, the target area is partitioned into non-overlapping regions. The partitioning method is versatile, allowing for options such as grid mapping, road networks, or locations like train/bus/bike stations. Given the inflow and outflow data up to time t−1, ST-SAM can capture the underlying, intrinsic and complicated spatial-temporal correlations and forecast the inflow and outflow of each region at time t.

ST-SAM first employs a novel region embedding layer to encode historical traffic data and learn embedding at a specific time for all the regions from their historical traffic data, so that the joint spatial-temporal correlation could be considered. Given that, it uses a spatial-temporal dependency learning module to capture the correlation between any two regions. Instead of only focusing on the local spatial correlation, ST-SAM considers correlations for both nearby and remote regions by evaluating correlations of any two regions. In contrast to previous approaches, the fusion of spatial, temporal, and flow information from preceding time slots within the region embedding enables the holistic consideration of spatial-temporal dependencies in the learning module. This comprehensive integration results in more precise predictions, as the interplay between spatial and temporal factors is effectively captured. Based on that, it uses a forecasting layer to predict the traffic data for regions.

We assess the effectiveness of our model using two traffic datasets. The results demonstrate that ST-SAM significantly outperforms the state-of-the-art methods, showcasing a notable enhancement in RMSE and MAPE metrics by approximately 15%. ST-SAM is much more efficient, with an improvement of up to 50 times on model training time.

## II. RELATED WORKS

Some early works propose using conventional time series analysis approaches (e.g., ARIMA) or machine learning models [7,8] for traffic forecasting. Although commendable, these approaches primarily concentrate on the temporal dependencies within traffic data, overlooking the crucial aspect of spatial dependency on other regions.

Recently, deep learning techniques have been utilized to capture spatial and temporal correlations in traffic forecasting. Convolutional Neural Networks (CNN) have been a prevalent choice in various studies to capture correlations among spatially adjacent regions by dividing a city into interconnected yet non-overlapping grids [3,4,9]. Some works use Graph Convolutional Networks (GCN) to depict spatial correlations, treating a city as a graph structure [10,11,12]. However, these approaches tend to focus on capturing local dependency on nearby regions, often missing out on distant spatial dependencies. In contrast, ST-SAM employs data-driven techniques to assess the interdependency between any two regions, considering both near and distant dependencies. Moreover, in comparison to CNN and GCN methods, ST-SAM offers greater generality and flexibility, accommodating various city representations such as grid maps, graphs, or other partitioning schemes.

Recurrent Neural Networks (RNN) have traditionally been harnessed to capture temporal correlations for traffic forecasting [7,12]. Nonetheless, RNN architectures are computationally demanding. Given that recurrent networks produce a series of hidden states based on prior hidden states and input at each step, they typically pose challenges for parallel training [13]. In contrast to RNN-based frameworks, the self-attention mechanism employed in our proposed model can effectively model long-term temporal dependencies while enabling parallel computation.

Furthermore, the majority of existing research addresses spatial and temporal dependencies sequentially, often neglecting their combined impact. In contrast, our proposed model integrates spatial and temporal dependencies jointly, allowing for a more comprehensive capture of their joint effects.

## III. METHODOLOGY

We overview the proposed model in Fig. 1. There are two crucial modules in ST-SAM. To predict the traffic data of individual regions at time t + 1, ST-SAM first uses the region representation learning module to generate embedding of an individual region at time slot t, by fusing the spatial information, temporal information, and flow information until time t. Based on that, it employs the spatial-temporal dependency learning module to update the region embedding by capturing the joint spatial-temporal dependency between any two regions. After that, a forecasting layer is leveraged to predict the inflow and outflow at time $t + 1$ for any individual region based on the updated region embedding.

## A. Region Representation Learning Module

Given the inflow data $\mathcal{I}^i = \{I_1^i, I_2^i, \cdots, I_t^i\}$ and outflow data $\mathcal{O}^i = \{O_1^i, O_2^i, \cdots, O_t^i\}$ of a region $r_i$ until time t, we use an inflow embedding layer and an outflow embedding layer to learn the inflow and outflow representation $\mathcal{F}_{in}^{(i,t)}$ and $\mathcal{F}_{out}^{(i,t)}$ for $r_i$ at time t:

$$\mathcal{F}_{in}^{(i,t)} = \mathcal{I}_{t-k+1:t}^i \times W_1 + b_1, \quad (1)$$

$$\mathcal{F}_{out}^{(i,t)} = \mathcal{O}_{t-k+1:t}^i \times W_2 + b_2, \quad (2)$$

where $\mathcal{I}_{t-k+1:t}^i$ and $\mathcal{O}_{t-k+1:t}^i$ represent the inflow and outflow data from time slots $t-k+1$ to t, and $W_1$, $W_2$, $b_1$, $b_2$ are learnable parameters.

Moreover, to consider the correlation between the inflow and outflow data, we propose using a flow correlation layer to fuse their correlation:

$$\hat{\mathcal{F}}^{(i,t)} = (\mathcal{F}_{in}^{(i,t)} \parallel \mathcal{F}_{out}^{(i,t)}) \times W_3 + b_3, \quad (3)$$

where $W_3$ and $b_3$ are trainable parameters, and $\parallel$ if the operation for concatenation.

To consider the spatial information of region $r_i$, we use a spatial embedding layer to extract latent spatial features from its one-hot representation:

$$\mathcal{S}_i = \mathcal{R}_i \times W_4 + b_4, \quad (4)$$

where $\mathcal{R}_i$ is the one-hot vector with the $i$-th elements as 1 and 0 otherwise, and $W_4$, $b_4$ are learnable parameters.

Furthermore, we extract temporal features using the temporal embedding layer:

$$\mathcal{T}_t = \mathcal{H}_t \times W_5 + b_5 \quad (5)$$

where $\mathcal{H}_t$ is the one-hot vector for time t, and $W_5$, $b_5$ are learnable parameters.

Based on the representation of spatial, temporal and flow information, we further fuse them into a region embedding using a fusion layer:

$$\mathcal{R}_i^t = (\hat{\mathcal{F}}^{(i,t)} \oplus \mathcal{S}_i \oplus \mathcal{T}_t) \times W_6 + b_6, \quad (6)$$

where $\mathcal{R}_i^t$ is the region embedding for region $r_i$ at time t, $W_6$ and $b_6$ are trainable parameters, and $\oplus$ is the addition operation.

## B. Spatial-Temporal Dependency Learning Module

Based on the results of the region representation learning module, we propose the spatial-temporal dependency learning module with self-attention mechanism to update the region embedding by considering the spatial and temporal dependency between regions.

Our spatial-temporal dependency learning module uses a multi-head mechanism so as to effectively capture diverse types of correlations. For the m-th head, the dependency on a region $r_j$ for a region $r_i$ is calculated as

$$A_{i,j}^t = \frac{(\mathcal{R}_i^t \times W_Q^m) \times (\mathcal{R}_j^t \times W_K^m)^T}{\sqrt{d}}, \quad (7)$$

where $W_Q^m$ and $W_K^m$ are trainable parameters, d is a hyper parameter to denote the size of the embedding.

The new embedding of a region $r_i$ for the m-th head is then updated as

$$\mathcal{R}_i^{(t,m)} = \sum_{j=1}^n \frac{A_{i,j}^t}{\sum_{u=1}^n \exp(A_{i,u}^t)} \times \mathcal{R}_j^t. \quad (8)$$

Based on that, we fuse the results from different heads as

$$\hat{\mathcal{R}}_i^t = (\mathcal{R}_i^{(t,1)} \parallel \mathcal{R}_i^{(t,2)} \parallel \cdots \parallel \mathcal{R}_i^{(t,M)}) \times W_7, \quad (9)$$

where M represents head number, $\parallel$ denotes the concatenation operation of two embedding, and $W_7$ is a trainable parameter.

The output is then subsequently fed into the fully connected feed-forward network. Consistent with the Transformer architecture, we incorporate the Add & Norm block along with residual connections, illustrated in Fig 1.

## C. Forecasting Layer

Given the region embedding with spatial-temporal dependency of region $r_i$ at time t, the forecasting layer uses fully connected neural networks to predict the inflow and outflow for $r_i$ at $t+1$ simultaneously:

$$(\hat{I}_{t+1}^i, \hat{O}_{t+1}^i) = RELU(\hat{\mathcal{R}}_i^t \times W_{P_1} + b_{P_1}) \times W_{P_2} + b_{P_2}, \quad (10)$$

where $W_{P_1}$, $W_{P_2}$, $b_{P_1}$, and $b_{P_2}$ are learnable parameters, and $RELU(\cdot)$ is the activation function. In our work, we use the following loss function to train our model:

$$LOSS = \sqrt{\frac{\sum_{i=1}^n (I_{t+1}^i - \hat{I}_{t+1}^i)^2 + \sum_{i=1}^n (O_{t+1}^i - \hat{O}_{t+1}^i)^2}{2n}} \quad (11)$$

where $n$ represents the total number of regions.

## IV. ILLUSTRATIVE EXPERIMENTAL RESULTS

### A. Experiemntal Settings

We extensively experiment with two real-world open datasets, namely the NYC-Taxi dataset and the NYC-Bike dataset [3]. The NYC-Taxi dataset contains taxi trips in New York city from January 1, 2015 to March 1, 2015, and the NYC-Bike dataset contains bike trips at New York city from July 1, 2016 to August 29, 2016. For both datasets, we divided the city into 200 non-overlap grids and extracts the inflow and outflow data for each grid at a time interval of 30 minutes. Our experimental design involved utilizing the initial 40 days of data as training data, with the subsequent days earmarked for testing purposes. During the training phase, 80% of the training data was assigned for model training, with the remaining 20% set aside for validation purposes.

We have implemented our model using PyTorch with the following hyperparameters: The number of heads is 4, the feedforward dimension is 128, and the embedding dimension, denoted as d, is 64. For the NYC-Taxi dataset, we set the historical time slot length k to 5, and for the NYC-Bike dataset, it is set to 4. The dropout rate is set at 0.1, the learning rate is 0.001, and the batch size is 64. Training was conducted on a machine equipped with an RTX 3080Ti GPU.

We assess our model against the following approaches using Root Mean Squared Errors (RMSE) and Mean Average Percentage Error (MAPE):

- ARIMA: It is a popular time series analysis method used for forecasting.
- XGBoost [14]: It is a widely used supervised machine learning model.
- STDN [3]: It predicts traffic flow using the combination of CNN and LSTM.
- ASTGCN [4]: It predicts traffic flow using an attention-based spatial-temporal graph convolution network.
- DSAN [6]: It learns the spatial-temporal correlation for traffic forecasting based on Transformer model.
- STGODE [11]: It predicts traffic flow using a spatial-temporal graph ordinary differential equation network based on a spatial graph and a semantic graph.
- ST-SSL [12]: It is a novel spatial-temporal self-supervised learning framework with spatial and temporal convolution for traffic forecasting.

*B. Accuracy*

The performance comparison between our approach and the state-of-the-art approaches on the NYC-Taxi and NYC-Bike is presented in Table I and II, respectively. Our proposed model demonstrates superior performance across both RMSE and MAPE metrics when compared to existing methods. Specifically, it achieves an average improvement of 15.65% in RMSE and 17.05% in MAPE. These substantial improvements demonstrate the efficacy of our model in effectively capturing the complicated spatial-temporal dependencies among regions.

*C. Training Efficiency*

In addition to prediction accuracy, we also evaluate the training efficiency of our model. Table III presents a comparison with state-of-the-art methods regarding the average time per epoch and total training time. Our model is significantly efficient when compared to other deep learning models, showcasing an average improvement of 32 times in total training time. The results show the substantial efficiency of our model for traffic forecasting. The reason is that our model entirely relies on self-attention to capture dependency, and self-attention can parallelize computations more efficiently, as each element can be processed independently.

Table I. Prediction Performance on NYC-Taxi dataset

| Approach | Inflow | | Outflow | |
| --- | --- | --- | --- | --- |
| | RMSE | MAPE | RMSE | MAPE |
| ARIMA | 27.25 | 21.14% | 36.53 | 22.21% |
| XGBoost | 21.72 | 18.70% | 26.07 | 19.35% |
| ASTGCN | 22.05 | 20.25% | 26.10 | 20.30% |
| STDN | 19.05 | 16.25% | 24.10 | 16.30% |
| DSAN | 18.32 | 16.07% | 24.27 | 17.70% |
| STGODE | 21.46 | 19.22% | 27.24 | 19.30% |
| ST-SSL | 18.69 | 16.21% | 23.62 | 16.04% |
| Ours | **18.23** | **15.53%** | **22.79** | **15.37%** |

TABLE II. PREDICTION PERFORMANCE ON NYC-BIKE DATASET

| Approach | Inflow | | Outflow | |
| --- | --- | --- | --- | --- |
| | RMSE | MAPE | RMSE | MAPE |
| ARIMA | 11.25 | 25.79% | 11.53 | 26.35% |
| XGBoost | 8.94 | 22.54% | 9.57 | 23.52% |
| ASTGCN | 9.25 | 22.25% | 9.34 | 23.13% |
| STDN | 8.15 | 20.87% | 8.85 | 21.84% |
| DSAN | 7.97 | 20.23% | 10.07 | 23.92% |
| STGODE | 8.58 | 23.33% | 9.23 | 23.99% |
| ST-SSL | 7.92 | 20.06% | 8.77 | 21.37% |
| Ours | **7.89** | **19.36%** | **8.09** | **19.59%** |

TABLE III. TRAINING EFFICIENCY OF DIFFERENT APPROACHES

| Approach | NYC-TAXI | | NYC-BIKE | |
| --- | --- | --- | --- | --- |
| | Average time per epoch (s) | Total time (s) | Training time/epoch (s) | Total time (s) |
| XGBoost | - | 313.67 | - | 266.02 |
| ASTGCN | 25.31 | 6272.88 | 25.68 | 5084.64 |
| STDN | 445.47 | 34746.66 | 480.21 | 23066.43 |
| DSAN | 386.17 | 29390.75 | 434.03 | 26476.20 |
| STGODE | 18.53 | 3423.48 | 18.76 | 3752.89 |
| ST-SSL | 3.92 | 486.08 | 3.89 | 420.40 |
| Ours | 3.19 | 440.03 | 3.17 | 357.74 |

V. CONCLUSIONS

We study the problem of traffic forecasting, and propose ST-SAM to capture the complicated and joint spatial-temporal dependency. We first propose a region representation learning module to learn the time-specific embedding of regions by encoding the spatial, temporal, and historical traffic information. We then propose a spatial-temporal dependency learning module based on self-attention mechanism to capture the joint dependency between any two regions, no matter they are nearby or distant. We evaluate ST-SAM through extensive experiments using a taxi dataset and a bike dataset. Our experimental results demonstrate that ST-SAM is significantly efficient and effective compared to existing approaches. Leveraging the self-attention mechanism, it significantly outperforms the state-of-the-art methods, achieving substantial improvements of up to 18% on

RMSE, 19% on MAPE, and a remarkable 32-fold reduction in training time.

## VI. ACKNOWLEDGEMENT

This research was funded by the Research Fund from Guangzhou Urban Planning & Design Survey Research Institute Co., Ltd. (RDI2230201023), Collaborative Innovation Center for Natural Resources Planning and Marine Technology of Guangzhou (No. 2023B04J0301), the Guangdong Enterprise Key Laboratory for Urban Sensing, Monitoring and Early Warning (No. 2020B12120219), and the National Key R&D Program of China (2022YFB3904105).